\documentclass[journal]{IEEEtran}
\usepackage{times}
\usepackage{epsfig}
\usepackage{graphicx}
\usepackage{amsmath}
\usepackage{amssymb}

\usepackage{subfig}
\usepackage{booktabs}
\usepackage{float}
\usepackage[ruled, linesnumbered]{algorithm2e} 
\usepackage{amsthm} 

\usepackage{makecell,rotating}
\usepackage{color,xcolor}
\usepackage{multirow}
\usepackage{capt-of,etoolbox}
\usepackage{caption}
\usepackage{mathtools}
\usepackage[colorlinks,linkcolor=blue]{hyperref}

\ifCLASSINFOpdf
\else
\fi
\begin{document}
%
\title{Arbitrary Virtual Try-On Network: Characteristics Preservation and Trade-off between Body and Clothing}
%
%
%

\author{Yu~Liu,~
        Mingbo~Zhao,~\IEEEmembership{Senior~Member,~IEEE,}
        Haijun~Zhang,~\IEEEmembership{Senior~Member,~IEEE,}
	     Zhao~Zhang,~\IEEEmembership{Senior~Member,~IEEE,}
        Shuicheng~Yan,~\IEEEmembership{Fellow,~IEEE}
\thanks{Manuscript received Feb. 19, 2021; revised August 26, 2021. This work is partially supported by National Key Research and Development Program of China (2019YFC1521300), supported by National Natural Science Foundation of China (61971121, 61672365) and also supported by the Fundamental Research Funds for the Central Universities of China (JZ2019HGPA0102).}
\thanks{Y. Liu and M. Zhao are with the School of Information Science and Technology,
Donghua University, Shanghai, China. (e-mail: 2191408@mail.dhu.edu.cn, mzhao4@dhu.edu.cn).}
\thanks{Haijun Zhang is with the Department of Computer Science, Harbin Institute of Technology, Shenzhen, Shenzhen 518055, China (e-mail: hjzhang@hit.edu.cn).}
\thanks{Zhao Zhang is with the Department of Computer Science, Hefei University of Technology, Hefei 230009, China (e-mail: cszzhang@gmail.com).}
\thanks{Mingliang Xu is with the School of Information Engineering, Zhengzhou University, Zhengzhou 450001, China (e-mail: iexumingliang@zzu.edu.cn).}
\thanks{Shuicheng Yan is with Qihoo 360, Beijing, China, and also with the National University of Singapore, Singapore 119077 (e-mail: eleyans@nus.edu.sg).}}

%
%

\markboth{Journal of \LaTeX\ Class Files,~Vol.~14, No.~8, August~2015}%
{Shell \MakeLowercase{\textit{et al.}}: Bare Demo of IEEEtran.cls for IEEE Journals}
%



\maketitle
\makeatletter
\DeclareRobustCommand\onedot{\futurelet\@let@token\@onedot}
\def\@onedot{\ifx\@let@token.\else.\null\fi\xspace}

\def\eg{\emph{e.g}\onedot} \def\Eg{\emph{E.g}\onedot}
\def\ie{\emph{i.e}\onedot} \def\Ie{\emph{I.e}\onedot}
\def\cf{\emph{c.f}\onedot} \def\Cf{\emph{C.f}\onedot}
\def\etc{\emph{etc}\onedot} \def\vs{\emph{vs}\onedot}
\def\wrt{w.r.t\onedot} \def\dof{d.o.f\onedot}
\def\etal{\emph{et al}\onedot}
\makeatother
\begin{abstract}
    Deep learning based virtual try-on system has achieved some encouraging progress recently, but there still remain several big challenges that need to be solved, such as trying on arbitrary clothes of all types, trying on the clothes from one category to another and generating image-realistic results with few artifacts. To handle this issue, we in this paper first collect a new dataset with all types of clothes, \ie tops, bottoms, and whole clothes, each one has multiple categories with rich information of clothing characteristics such as patterns, logos, and other details. Based on this dataset, we then propose the Arbitrary Virtual Try-On Network (AVTON) that is utilized for all-type clothes, which can synthesize realistic try-on images by preserving and trading off characteristics of the target clothes and the reference person. Our approach includes three modules: 1) Limbs Prediction Module, which is utilized for predicting the human body parts by preserving the characteristics of the reference person. This is especially good for handling cross-category try-on task (\eg long sleeves \(\leftrightarrow\) short sleeves or long pants \(\leftrightarrow\) skirts, \etc), where the exposed arms or legs with the skin colors and details can be reasonably predicted; 2) Improved Geometric Matching Module, which is designed to warp clothes according to the geometry of the target person. We improve the TPS based warping method with a compactly supported radial function (Wendland's \(\Psi\)-function); 3) Trade-Off Fusion Module, which is to trade off the characteristics of the warped clothes and the reference person. This module is to make the generated try-on images look more natural and realistic based on a fine-tune symmetry of the network structure. Extensive simulations are conducted and our approach can achieve better performance compared with the state-of-the-art virtual try-on methods.
\end{abstract}
\section{Introduction}
In modern society, clothing plays an important role in human daily life, as proper outfits can enhance peoples’ beauty and personal quality \cite{entwistle2015fashioned}. But in practice, people tend to make the final purchase decision before trying on the clothes. On the other hand, huge amounts of clothing data emerge on the Internet or social media during the past few years due to the development of the clothing e-commerce platform. Therefore, it is impossible and unfeasible for people to try on all the clothes especially on the website. Therefore, to meet the shopping needs of consumers \cite{haubl2000consumer}, it is necessary for researchers to conducted considerable researches on image-based virtual try-on technology \cite{hauswiesner2013virtual, iccv2017fashiongan}, which can easily help people to achieve image-realistic try-on effects on website or smartphone app by transforming the target clothing outfit onto a certain reference person.

During the past few years, computer vision technology has been widely utilized in the extensive applications of artificial fashion. These applications include clothes detection \cite{liuLQWTcvpr16DeepFashion, ge2019deepfashion2}, clothes parsing \cite{li2019self, liang2015deep, li2017multiple, gong2017look}, clothes attributions and categories recognition \cite{wang2018toward, liuLQWTcvpr16DeepFashion, ge2019deepfashion2}, clothes collocation \cite{ iwata2011fashion, zhao2020end, veit2015learning, he2016learning, shih2018compatibility, li2017mining, han2017learning, cui2019dressing}, clothes image retrieval \cite{zeng2020tilegan, liu2018deep, liuYLWTeccv16FashionLandmark, guo2019fashion, lin2020fashion, hosseinzadeh2020composed, jandial2020trace} and fashion edit \cite{hsiao2019fashion++,dong2019fashion}. These applications are all merited from recently developed technology, namely deep learning due to its powerful feature extraction ability to capture the rich mid-level image representations. Motivated by this technology, the deep learning based virtual try-on methods have been extensively studied and achieved considerable results recently. These methods have adopted the concept of GAN \cite{goodfellow2020generative} and focus on generating the synthetic and realistic try-on images by preserving the characteristics of a clothing image (\eg, texture, logo, and embroidery) as well as warping it to an arbitrary human pose. Typical methods for virtual try-on include VITON \cite{han2018viton}, CP-VTON \cite{wang2018toward}, VTNFP \cite{yu2019vtnfp}, ACGPN \cite{yang2020towards}, \etal. \cite{pandey2020poly, neuberger2020image, raj2018swapnet,pons2017clothcap, mo2018instagan, guan2012drape, iccv2017fashiongan, han2019compatible}. Extensive simulation results show the effectiveness of the proposed work. 
\begin{figure*}
\centering
\includegraphics[width=1\textwidth]{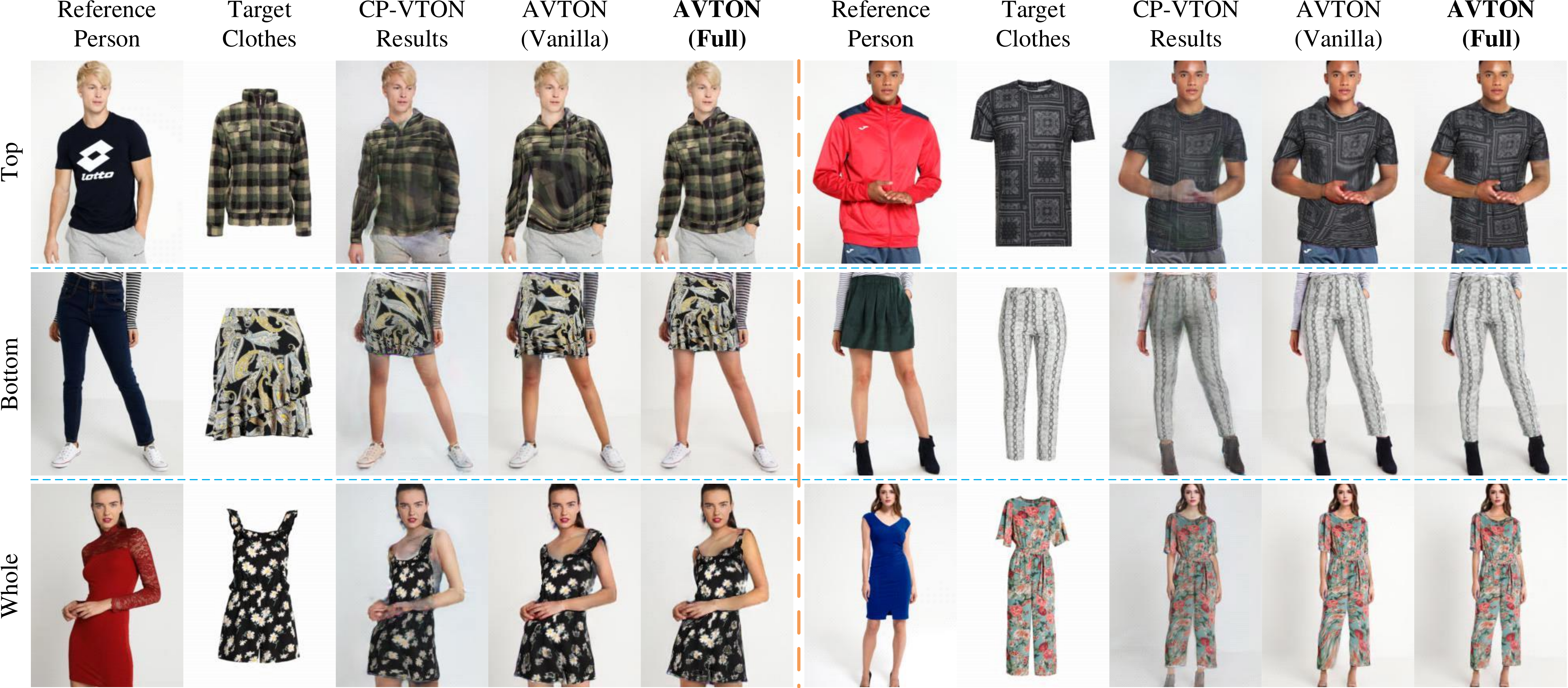}
		\captionof{figure}{We propose AVTON that is trained with an \emph{all-type clothing dataset}. It can be adapted to all-type clothing try-on task and get image-realistic results. The types of clothes are divided into the \emph{top, bottom, and whole}, and we apply CP-VTON \cite{wang2018toward} and AVTON on them. The CP-VTON is retained with the all-type clothing dataset, and the AVTON (Vanilla) indicates the AVTON trained without \emph{LPM and Wendland's  \(\Psi\)-function} \cite{wendland1995piecewise}, while the AVTON (Full) is just the opposite.}
\label{fig: beginning}
\end{figure*}

Although these researches have made some progress, it remains some ongoing challenges and limitations: 1) most benchmark datasets utilized for training virtual try-on methods mainly contain top clothes. Therefore, current related works can only focus on top clothing try-on task due to the lack of diversity in the dataset. For example, the VITON dataset \cite{han2018viton} is a benchmark dataset used in most prior arts. It only contains frontal-view woman and top clothing image pairs. As a result, the trained model can only handle top clothing try-on task while cannot be adapted to work on the other bottom or whole clothing try-on task. Therefore, how to collect a diverse try-on dataset with tops, bottoms, and whole clothes and develop a method for all-type clothing try-on task is necessary and important; 2) cross-category try-on task(\eg long sleeves\(\leftrightarrow\)short sleeves or long pants\(\leftrightarrow\)skirts, \etc) is another challenge in the virtual try-on system. A case in point is that when people aim to try on from long sleeves to short sleeves, some parts of people’s arms will be exposed. Therefore, it is necessary to preserve the characteristics of the reference person and predict such an exposed human body when generating the image-realistic try-on results. However, most current methods \cite{han2018viton,wang2018toward,pandey2020poly,yu2019vtnfp} mainly focus on try-on with one category by preserving the characteristics of clothes but do not consider the limb prediction when handling try-on with different categories. As a result, some methods may cause problems, \eg, the limb is covered by clothes, the color of the skin is wrongly painted and the detailed hands cannot be generated; 3) current methods are not good at trading off characteristics of the warped clothes and the reference person, such as ACGPN. Although it can preserve the characteristics of the warped clothes and reference person as much as possible, its generated images are not realistic enough, such as some artifacts near the neck regions. The reason for these results is that the final fusion module prefers to preserve the warped clothing characteristics and cannot correct these errors.

In this paper, we propose a new image-based virtual try-on method, called AVTON, to address the issues mentioned above: 1) In order to handle the all-type clothing try-on task and increase the diversity of the dataset, we collect a new dataset from the \href{https://www.zalando.co.uk}{Zalando}. As shown in Fig. \ref{fig: beginning}, the example in the first row represents the top clothing try-on task. This type of data is similar to the VITON dataset, but it involves more rich characteristics, such as human genders, shapes, and poses, and especially more categories in top clothing images. The second row provides the images for the bottom clothing try-on task, where the bottom clothing images include trousers, shorts, and skirts, \etc, with rich textures, logos, and other details. The last row provides examples for the whole clothing try-on task, where this type of human image generally contains the entire human body. Based on such a diverse dataset, we then develop the proposed method, which can naturally handle try-on task for all-type clothes; 2) We propose a new image-based virtual try-on method, called AVTON, to achieve arbitrary clothing try-on and image-realistic results. The proposed method contains three modules: a) Limbs Prediction Module, which is developed for predicting limbs, and keep the head and the non-target human body parts, to preserve the characteristics of the reference person. This module is especially suitable for handling cross-category try-on task, such as long sleeves \(\leftrightarrow\) short sleeves or long pants \(\leftrightarrow\) skirts, \etc, where the exposed arms or legs (including their skin colors and details) can be reasonably predicted. This is good to help the try-on system for formulating a realistic result in the following modules; b) Improved Geometric Matching Module, which is designed to warp clothes according to the geometry of the reference person. By carefully analyzing the basic concept of Thin-Plate Spline (TPS) \cite{duchon1977splines} based methods, we argue that the selection of radial basis function is a key point to affect the performance of image warping. Motivated by this end, we then have adopted Wendland's \(\Psi\)-function \cite{wendland1995piecewise} as the compactly supported radial basis function. Both theoretical analysis and simulation have verified that the proposed method is able to characterize the local geometrical structure of images, which is good for warping the clothes image especially with the complex texture; c) Trade-Off Fusion Module, which is to trade off the characteristics of the warped clothes and the reference person, this module is to make the generated try-on images looks more natural and realistic based on a fine-tune symmetry of the network structure (a pair of UNet \cite{ronneberger2015u}). Experiments show that AVTON significantly outperforms the state-of-the-art methods for virtual try-on \cite{han2018viton,wang2018toward,yu2019vtnfp,yang2020towards}, and can generate realistic try-on images in all-type clothing try-on task (Fig. \ref{fig: beginning}).

The main contributions of this paper are summarized as follows:
\begin{enumerate}
\item We collected a diverse dataset for handling the all-type clothing try-on task, which contains all types of clothes. Benefiting from this dataset, we are able to design an arbitrary virtual try-on method;
\item In order to handle the cross-category try-on task, we propose a new limbs prediction module to preliminarily predict limbs and preserve the characteristics of the reference person. We show that it is necessary for the cross-category try-on task;
\item We for the first time use the \(\Psi\)-function of Wendland to warp clothes, which is compact support for the registration of images. It greatly improves the try-on quality in preserving the characteristics of the target clothes;
\item We present a novel fusion network that trades off the characteristics between the warped clothes and the reference person. By adjusting characteristics and correcting errors, the try-on images learned are more realistic;
\item We demonstrate that our proposed method can be applied to the all-type clothing try-on task and cross-category try-on task, and outperforms the state-of-the-art methods both qualitatively and quantitatively.
\end{enumerate}
\section{Related Work}
\textbf{Image Synthesis}. Generative Adversarial Network (GAN) \cite{goodfellow2020generative} is a popular and effective tool in image synthesis \cite{isola2017image,karras2017progressive,karras2019style,park2019semantic} and generally consists of a generator and a discriminator. The core idea of GAN is a zero-sum game, in which the generator aims to generate a realistic image and the discriminator is to distinguish it from the real one. The procedure is continuous until the discriminator is unable to judge whether the output result of the generator is true or not. By taking advantage of the GAN, researchers make great progress in image-to-image translation \cite{isola2017image,zhu2017unpaired,mo2018instagan}, photo inpainting \cite{han2019compatible}, clothing translation \cite{zeng2020tilegan, pons2017clothcap}, \etc. Extensive learning tasks verify the effectiveness of image synthesis.

\textbf{Fashion Analysis}. Benefit from the development and popularization of deep learning, fashion-related work has made great progress recently. Current works of artificial fashion mainly focus on clothing parsing \cite{li2019self, liang2015deep, li2017multiple, gong2017look}, clothing landmark detection \cite{liuLQWTcvpr16DeepFashion, ge2019deepfashion2}, clothes attributions and categories recognition \cite{wang2018toward, liuLQWTcvpr16DeepFashion, ge2019deepfashion2}, fashion retrieval \cite{zeng2020tilegan, liu2018deep, liuYLWTeccv16FashionLandmark, guo2019fashion, lin2020fashion, hosseinzadeh2020composed, jandial2020trace} and fashion recommendation \cite{iwata2011fashion, zhao2020end, veit2015learning, he2016learning, shih2018compatibility, li2017mining, han2017learning, cui2019dressing}. Additionally, the deep learning based virtual try-on task is one of the popular tasks of fashion analysis that has been widely studied and has made great achievements \cite{guan2012drape, han2018viton, wang2018toward, alldieck2018video, pons2017clothcap, yu2019vtnfp, mir2020learning, yang2020towards}.

\textbf{Virtual Try-On}. Virtual try-on belongs to fashion image synthesis, which can be roughly divided into two categories: 3D model-based methods \cite{alldieck2018video, pons2017clothcap, mir2020learning, guan2012drape} and 2D image-based methods \cite{han2018viton, wang2018toward, yu2019vtnfp, yang2020towards}. At present, both methods are based on deep learning and make great progress. However, the dataset on the 3D model-based methods is difficult to obtain, and these methods need huge computation due to the complexity of the model; On the other hand, 2D image-based methods does not require auxiliary methods to improve the performance of the network so that it is more easily to be trained. Among all deep learning based 2D methods, VITON \cite{han2018viton} and CP-VTON \cite{wang2018toward} are the two first works. The key idea of VITON is to exploit a Thin-Plate Spline (TPS) \cite{duchon1977splines} based method to warp the clothes images with texture mapped on it, while CP-VTON has extended VITON by developing neural network layers to learn the transformation parameters of TPS so that it is trainable and can achieve more correct alignment performances. Both two methods cannot achieve satisfactory results when postural changes and complexity texture occur. VTNFP \cite{yu2019vtnfp} and ACGPN \cite{yang2020towards} are another two popular methods that have improved the performance of the try-on task with more characteristics of clothes and human body preserved, but they still have problems that the generated images are not realistic enough, such as some artifacts near the neck regions, as the final fusion module in these methods prefers to preserve the warped clothing characteristics and cannot correct these errors. In addition, all these methods are trained via VITON dataset \cite{han2018viton} that only contains top clothes. As a result, these methods cannot handle arbitrary try-on tasks with all types of clothes. Another challenge is the cross-category try-on task. Though the work in \cite{mo2018instagan} has proposed a variant of CycleGAN \cite{zhu2017unpaired}, namely InstaGAN, to handle this task (long sleeves \(\leftrightarrow\) short sleeves, long pants \(\leftrightarrow\) skirts, \etc), the final try-on results cannot be controlled due to the strategy of CycleGAN. This paper aims to address the above problems by proposing a new method for all-type clothing and cross-category try-on tasks with more image-realistic performance.
\section{Arbitrary Virtual Try-On Network}
\begin{figure*}[h!]
\begin{center}
\includegraphics[width=0.9\textwidth]{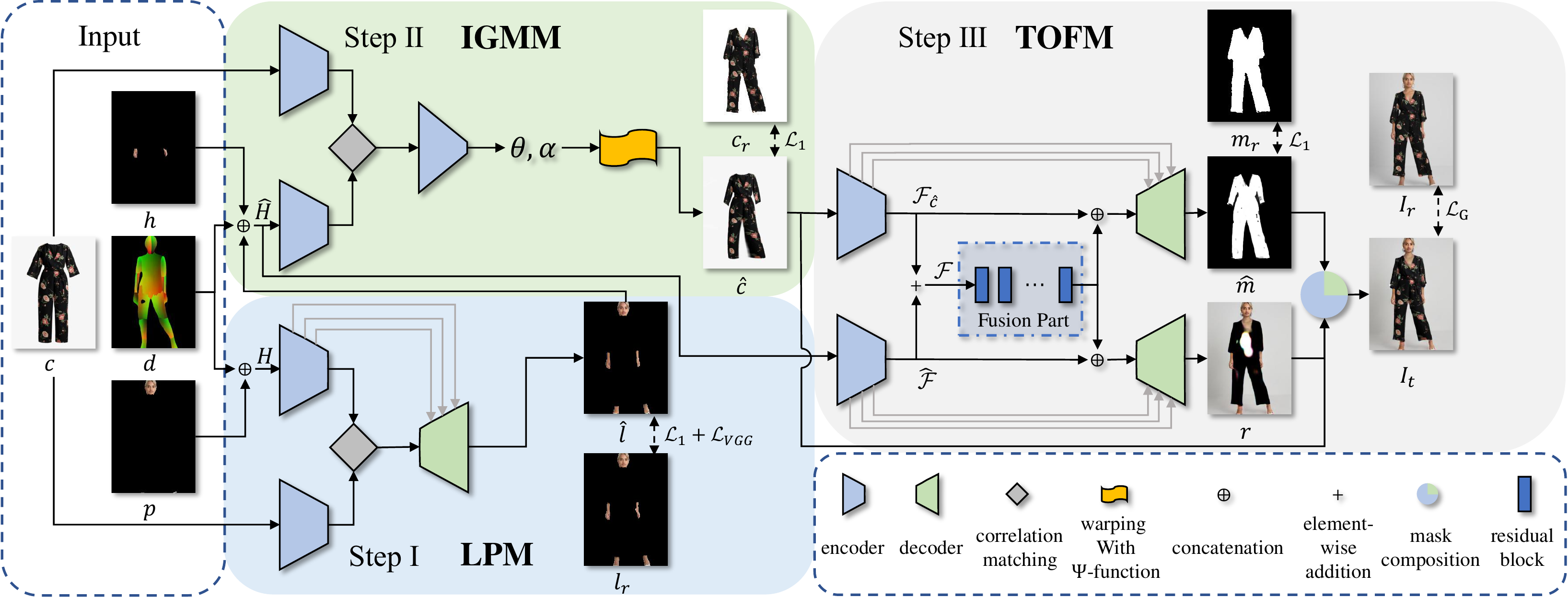}
\end{center}
   \caption{An overview of our AVTON. \textbf{Step \uppercase\expandafter{\romannumeral1}}: Limbs Prediction Module takes the target clothes \(c\) and the human body information \(H\) as the input to predict the exposed limbs and preserve the primary human body information, and output the predicted human body \(\hat{l}\); \textbf{Step \uppercase\expandafter{\romannumeral2}}: Improved Geometric Matching Module takes the target clothes \(c\) and the refined human body information \(\hat{H}\) as input, and output the warped clothes \(\hat{c}\); \textbf{Step \uppercase\expandafter{\romannumeral3}}: Trade-Off Fusion Module firstly takes the warped clothes \(\hat{c}\) and the refined human body information \(\hat{H}\) as the input to predict the composition mask \(\hat{m}\) and the rendered person \(r\), and then compose the outputs with the warped clothes \(\hat{c}\) to generate the try-on image \(I_t\).}
\label{fig:network}
\end{figure*}

Our goal is to learn an arbitrary virtual try-on model that can be adapt to all-type clothing and cross-category try-on tasks and generate more realistic try-on images than prior arts. The proposed AVTON contains three modules, as shown in Fig. \ref{fig:network}. First, the limbs Prediction Module (LPM) is utilized to predict the limbs, head, and non-target human body. This is especially useful for handling cross-category try-on task. Second, the Improved Geometric Matching Module (IGMM) uses Wendland's \(\Psi\)-function \cite{wendland1995piecewise} to improve the TPS based method \cite{duchon1977splines} for warping the clothes. Third, the Trade-Off Fusion Module (TOFM) takes the warped clothes and the human body information as input, and then generates composition mask and rendered person by a pair of UNet \cite{ronneberger2015u} and a fusion part. Especially, the TOFM takes advantage of GAN \cite{goodfellow2020generative}.
\subsection{Limbs Prediction Module (LPM)}
\label{sub: LPM}
The purpose of designing the Limbs Prediction Module (LPM) is to predict the exposed limbs and reserve the primary human body information (\ie, head, non-target human body parts). Most earlier methods generate exposed limbs in try-on steps but neglected primary human body information, which may lead to generate unreasonable color of skin and encountered occlusion issues. We thereby propose LPM to address these issues. 

Given a reference person image \(I_r\), LPM takes the target clothes \(c\) and the human body information (\ie, the pose and shape information \(d\) which is extracted from the reference person \(I_r\) by DensePose \cite{alp2018densepose}, the head and non-target human body parts information \(p\) in the reference person \(I_r\)) as inputs to predict exposed limbs \(\hat l\). In detail, the target clothes \(c\) and the human body information \(H=d\oplus p\) (\(\oplus\): concatenation) are firstly encoded as the input features. They are then formed as a single tensor by a correlation layer and input to the decoder. Finally, the exposed limbs are predicted by the decoder. The encoder and correlation layer are similar to CP-VTON's GMM step \cite{wang2018toward}, while the human body information \(d\)'s encoder-decoder layers are similarly to UNet \cite{ronneberger2015u} structure shown in Fig. \ref{fig:network}. All this leads to preserve the primary body information.

The LPM is trained under a combination of the pixel-wise L1 loss and VGG perceptual loss between predicted result \(\hat l\) and ground truth \(l_r\), where \(l_r\) includes head, non-target human body parts and exposed limbs in the reference person \(I_r\):
\begin{equation}
    \label{eq:LPM loss}
    \mathcal{L}_{\text{LPM}}=\lambda_{\text{L1}}\Vert \hat{l}-l_r\Vert_1+\lambda_{\text{vgg}}\mathcal{L}_{\text{VGG}}(\hat{l}, l_r)
\end{equation}
where
\begin{equation}
\label{equ: vgg loss}
\mathcal{L}_{\text{VGG}}(\hat{l}, l_r)=\sum_{i=1}^5\lambda_i\Vert \phi_i(\hat{l})-\phi_i(l_r)\Vert_1
\end{equation}
is the VGG perceptual loss, where \(\lambda_{\text{L1}}\) and \(\lambda_{\text{vgg}}\) are the trade-off parameters for two loss terms in Eq. \ref{eq:LPM loss}, which all set to 1 in our experiments, and \(\phi_i(l)\) denotes the feature map of limbs' image \(l\) of the {\it \(i\)-th} layer in the visual perception network \(\phi\), which is a VGG19 pre-trained on ImageNet.
\subsection{Improved Geometric Matching Module \\ (IGMM) }
\label{sub: IGMM}
The old way of warping clothes is based on Thin-Plate Splines (TPS) \cite{duchon1977splines} with \(r^2\log r\) as Radial Basis Functions (RBFs). This method yields minimal bending energy properties measured over the whole image. But since it is not a compactly supported RBFs, the deformation will cover the regions where all control points are located. It is advantageous for yielding an overall smooth deformation and preserving geometrical characteristics, but it is problematic when only a small part of the image is desired to be deformed. This will lead to unreasonable deformation when warps clothes. To address this issue,we in this work has adopted \(\Psi\)-function of Wendland \cite{wendland1995piecewise} as RBFs. As mentioned in \cite{fornefett1999elastic, fornefett2001radial}, it is a more compactly support for the registration of images so that the bending region can be narrowed down when minimizing the bending energy. Here, we first give its formulation as follows:
\begin{equation}
    \label{equ:wendland_0}
    \psi_{d,k}(r)\coloneqq I^k(1-r)_+^{\lfloor d/2\rfloor+k+1}(r)
\end{equation}
where
$$
(1-r)_+^v=
\begin{cases}
(1-r)^v& 0\le r<1\\
0& r\ge 1
\end{cases},
$$
$$
I\psi(r)\coloneqq \int_r^\infty t\psi(t)\text{d}t\quad r\ge 0.
$$
The equation also holds for different spatial supports \(\alpha\): \(\psi_\alpha(r)=\psi(r/\alpha)\). We apply the \(\psi_{\alpha,3,1}\)-function as RBFs to replace the TPS's RBFs:
\begin{equation}
\label{equ: wendland_1}
\psi_{\alpha,3,1}(r)=(1-r/\alpha)_+^4(4r/\alpha+1)
\end{equation}
where \(\alpha\) is a learnable parameter as same as the spatial transformation parameters \(\theta\). Inspired by CP-VTON \cite{wang2018toward}, we use the same structure to learn these parameters. As shown in Fig. \ref{fig:network}, our IGMM firstly extract high-level features of the target clothes \(c\) and the refined human body information \(\hat{H}=d\oplus \hat{l} \oplus h\) (\(h\) represents hand details in reference person \(I_r\)) respectively. Then a correlation layer to combine two features into as single tensor as input to the regression network that predicts \(\theta\) and \(\alpha\). Finally, a transformation \(\Psi_{\theta, \alpha}\) based on Eq. \ref{equ: wendland_1} for warping the target clothes \(c\) into the result \(\hat{c}=\Psi_{\theta, \alpha}(c)\).

To learn the module above, we make some derivations and experiments to study the size of spatial supports \(\alpha\), which shows a significant positive correlation between spatial warping range and \(\alpha\), and we concluded that the best \(\alpha\) should meet the condition: \(\alpha \ge D\), where \(D\) is the maximum distance among control points in Delaunay triangles \cite{delaunay1934bulletin}. In our experiments, we set \(D=\sqrt{a^2+b^2}\) (\(a\) and \(b\) is the vertical distance and the horizontal distance among nearest control points, respectively). Consequently, the final \(\alpha=\lambda_{\alpha}\hat{\alpha}+d\), where \(\hat{\alpha}\) is the sigmoid's output of the regression network and \(\lambda_{\alpha}\) is set to \(6\) in our experiments.

To train the module, we conducted it under the pixel-wise L1 loss between the warped clothes \(\hat{c}\) and ground truth \(c_r\), where \(c_r\) is the clothes worn on the reference person in \(I_r\):
\begin{equation}
    \label{equ:IGMM loss}
    \mathcal{L}_{\text{IGMM}}=\Vert \hat{c}-c_r\Vert_1=\Vert \Psi_{\theta,\alpha}(c)-c_r\Vert_1
\end{equation}
\begin{figure*}[h!]
\begin{center}
\includegraphics[width=0.8\textwidth]{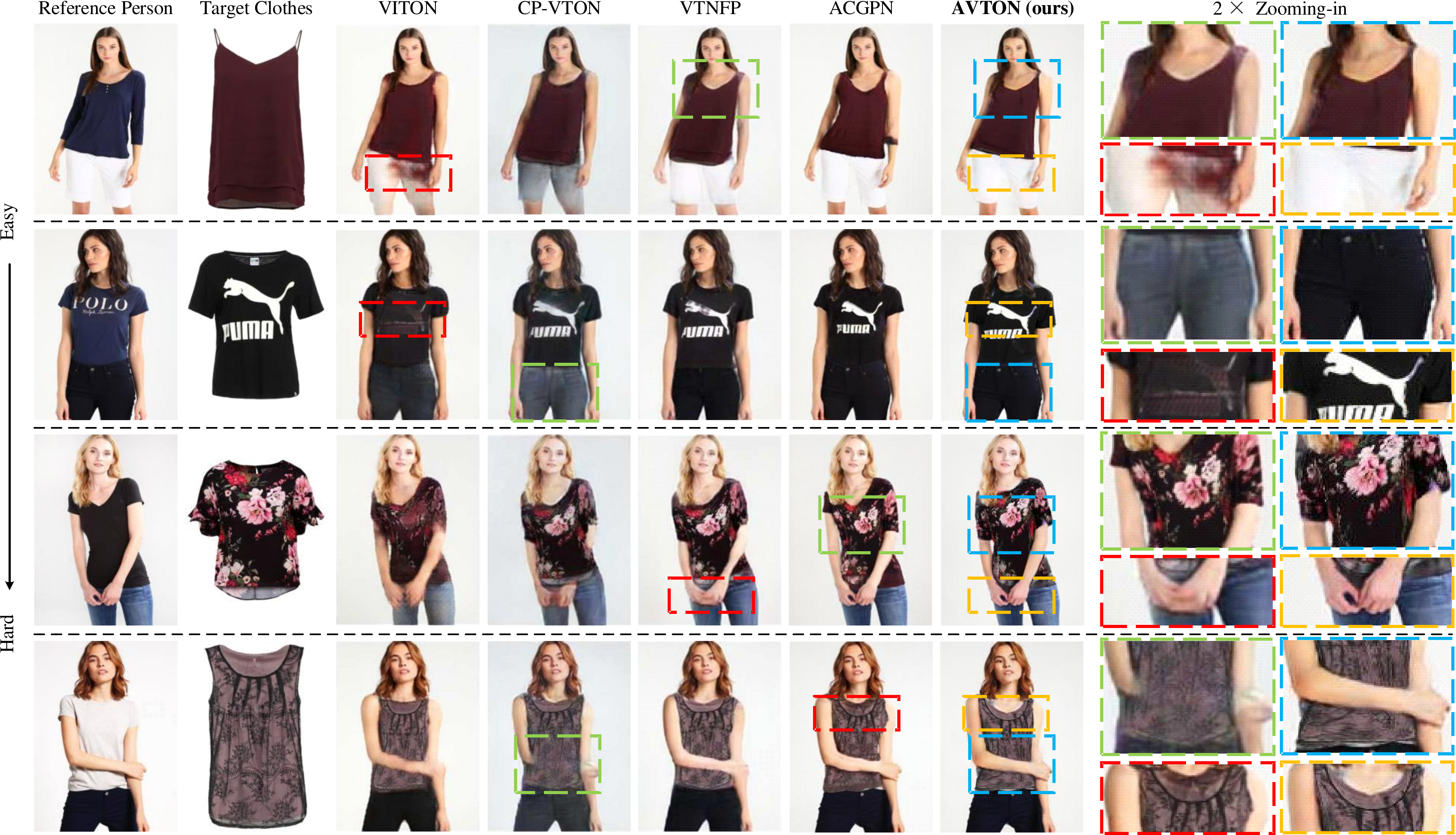}
\end{center}
   \captionof{figure}{\textbf{On the VITON-Dataset}. Qualitative comparisons of VITON \cite{han2018viton}, CP-VTON \cite{wang2018toward}, VTNFP \cite{yu2019vtnfp}, ACGPN \cite{yang2020towards} and AVTON in easy to hard levels (from top to bottom). Our method preserves more characteristics of the reference person with the LPM, and it also preserves more characteristics of the targets clothes with the IGMM. What is more, AVTON can generate more realistic try-on images with the TOFM, which is good at trading off characteristics of the warped clothes and the reference person.}
\label{fig:qualitative_results_0}
\end{figure*}
\subsection{Trade-Off Fusion Module (TOFM)}
\label{sub: TOFM}
Try-on image synthesis from the warped clothes and the reference person is a many-to-one mapping problem. It aims at not only preserving the characteristics of the warped clothes and the reference person, but also trading off them to make images realistic. One of the common methods \cite{han2018viton} is to produce a composition mask for fusing UNet \cite{ronneberger2015u} rendered person with warped clothes and finally to produce a refined result. Although it can refine the course try-on image, it lacks preserving characteristics of the warped clothes. Another common method \cite{wang2018toward} is to utilize an UNet to render a person image and predict a composition mask simultaneously, and then synthesizing the try-on image by fusing the rendered person and the warped clothes via the composition mask. But this way failed to preserve characteristics of reference person on account of using a single UNet structure, this structure prefers to preserve characteristics of the warped clothes. And other methods \cite{yu2019vtnfp, yang2020towards} have some problems with trading off characteristics between the warped clothes and the reference person (\eg, artifacts near the neck regions).

In this paper, we adopt a GAN \cite{goodfellow2020generative} based method for generating the realistic try-on image. In detail, we formulate the generator \(G\) in GAN by a pair of UNet, where the inputs are the warped clothes \(\hat{c}\) and refined human body information \(\hat{H}\), while the outputs are the generated composition mask \(\hat{m}\) and rendered person \(r\). However, making the characteristics of the warped clothes and the reference person contribute equally when generating the try-on image is still problematic, as this will cause occlusion and artifact problems. To make images more realistic, we add a fusion part in our try-on module (Fig. \ref{fig:network} residual blocks) for handling the above problems. Specifically, in the proposed module, the warped clothes' features \(\mathcal{F}_{\hat{c}}\) and the refined human body features are firstly summed element-wisely, so that the fused features can be obtained, i.e. \(\mathcal{F}\) (\(\mathcal{F}_{\hat{c}}+\mathcal{\hat{F}}\rightarrow\mathcal{F}\)). Then the fused features \(\mathcal{F}\) are concatenated with \(\mathcal{F}_{\hat{c}}\) and \(\mathcal{\hat{F}}\), and decoded respectively to get predicted mask \(\hat{m}\) and predicted rendered person \(r\). Finally, similarly to CP-VTON \cite{wang2018toward}, \(\hat{c}\) and \(r\) are fused together using \(\hat{m}\) to synthesize try-on image \(I_t\):
\begin{equation}
    \label{equ:TOFM try-on}
    I_t=\hat{m}\odot \hat{c}+(1-\hat{m})\odot r
\end{equation}
where \(\odot\) represents element-wise matrix multiplication. Additionally, we use the multi-scale discriminators \(D\) that is similar to pix2pixHD \cite{wang2018high}.

At the training phase, the generator's loss is the combination loss scheme of CP-VTON and pix2pixHD, it includes L1 loss, VGG perceptual loss (Eq. \ref{equ: vgg loss}) and LSGAN loss:
\begin{equation}
    \label{equ:TOFM g loss}
    \begin{split}
    \mathcal{L}_{G}=&\lambda_{L1}\Vert I_t-I_g\Vert_1+\lambda_{vgg}\mathcal{L}_{\text{VGG}}(I_t, I_g)+\\
    &\lambda_{mask}\Vert \hat{m}-m_r\Vert_1+\lambda_{lsgan}(D(\hat{c},\hat{H},I_t)-1)^2
    \end{split}
\end{equation}
while the discirminator's loss is:
\begin{equation}
    \label{equ:TOFM d loss}
    \begin{split}
    \mathcal{L}_{D}=&((D(\hat{c},\hat{H},I_t))^2+(D(\hat{c},\hat{H},I_g)-1)^2) / 2
    \end{split}
\end{equation}
where \(I_g\) is the ground truth image, and \(I_g=I_r\) in training stage, \(m_r\) is the mask of \(c_r\). In our experiments, we set \(\lambda_{L1}\), \(\lambda_{vgg}\) and \(\lambda_{mask}\) to \(10\), while set \(\lambda_{lsgan}\) to \(1\).
\section{Experiments}
\subsection{Datasets}
Experiments are conducted on two datasets:  the VITON dataset \cite{han2018viton} that is used in VITON and CP-VTON \etal \cite{han2018viton, wang2018toward, yu2019vtnfp, yang2020towards}, and the self-collected Zalando dataset. In this paper we will call them \textbf{VITON-Dateset} and \textbf{Zalando-Dataset} respectively.

\textbf{VITON-Dataset}. It contains 16253 frontal-view woman and top clothing image pairs, which is split into a training set and a testing set with 14221 and 2032 pairs respectively. To further research and comparison, we use the method of ACGPN \cite{yang2020towards} to score the complexity of each woman image and divide VITON-Dataset into Easy, Medium, and Hard levels.

\textbf{Zalando-Dataset}. It contains 34928 frontal-view human (include man and woman) and clothing (include top, bottom, and whole) image pairs, it was split into a training set and testing set with 32746 and 2182 respectively. The training set contains 19185 tops, 10587 bottoms, and 2974 whole clothes, and the testing set contains 1310 tops, 692 bottoms, and 180 whole clothes. We also score the complexity of each human image except the lower human body image.

Note that we use the DensePose \cite{alp2018densepose} to extract the pose and shape information, and we use the SCHP \cite{li2019self} to obtain the human parses for the head, hand detail, and non-target human body parts.
\subsection{Implementation Details}
The experiments are conducted on VITON-Dataset and Zalando-Dataset respectively, and the results totally independent of each other.

\textbf{Training}. Followed by steps in Fig. \ref{fig:network}, we first train the LPM and then use the LPM's trained results to train IGMM, finally train the TOFM with the trained results of LPM and IGMM. On the VITON-Dataset training setup, each module is trained for 400K steps with batch size 4, while on the Zalando-Dataset training setup, each module is trained for 800K steps with batch size 4. Both training setups use Adam optimizer with \(\beta_1=0.5\) and \(\beta_2=0.999\), and learning rate is fixed at 0.0001. Additionally, the resolution for all input and output images is \(256\times 192\), and we use one NVIDIA 2080Ti GPU in our experiments.

\begin{figure*}[h!]
\begin{center}
\includegraphics[width=0.9\textwidth]{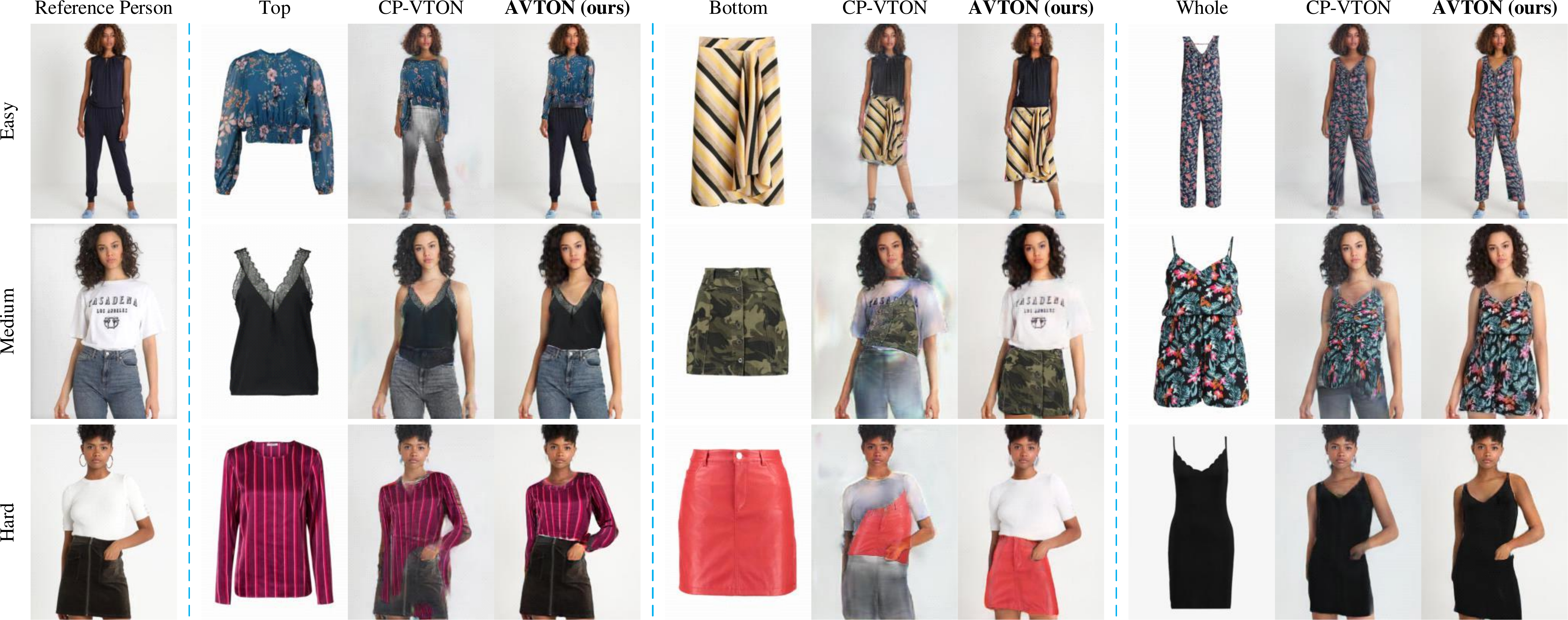}
\end{center}
   \captionof{figure}{\textbf{On the Zalando-Dataset}. Qualitative comparisons of CP-VTON and AVTON in different types of clothes. Our AVTON adapts successfully in all-type clothing try-on task.}
\label{fig:qualitative_results_1}
\end{figure*}

\textbf{Testing}. We use the same steps as the training stage to test modules. On the VITON-Dataset testing stage, we test our modules in easy, medium, and hard cases respectively, and evaluate the results qualitatively and quantitatively. On the Zalando-Dataset testing process, we evaluate the qualitative and quantitative results in the top, bottom, and whole clothing cases respectively, and perform ablation experiments. Note that we also analyze the qualitative results in the easy, medium, and hard cases respectively (there are complexity labels in Fig. \ref{fig:qualitative_results_1} and Fig. \ref{fig:qualitative_results_2}).

\subsection{Qualitative Results}
\label{sub: qualitative results}
\textbf{On the VITON-Dataset}. Fig. \ref{fig:qualitative_results_0} shows visual comparison of our proposed method with VITON \cite{han2018viton}, CP-VTON \cite{wang2018toward}, VTNFP \cite{yu2019vtnfp} and ACGPN \cite{yang2020towards} in different complexity levels. To save a lot of work to reproduce them (VTNFP has no official code), we refer to the results from the paper of ACGPN. Note we validated the official code of ACGPN qualitatively and quantitatively and got the same results with ACGPN.  

Among these results, VITON is the worst method in preserving characteristics, such as the blurred hands (\ie the red box of the first row), the disappeared pattern (\ie the red box of the second row), and so on. CP-VTON has a slight improvement on characteristics-preserving, but it still failed to keep the non-target parts (\ie the green box of the second row) and deal with large occlusions (\ie the green box of the fourth row). In a word, VITON and CP-VTON are bad at preserving characteristics of target clothes and reference person (mentioned in Section \ref{sub: LPM} and Section \ref{sub: IGMM}). 

In comparison to VITON and CP-VTON, VTNFP preserves more characteristics on account of using segmentation representation to preserve the non-target parts, but it does not contain enough details. For example, the thin shoulder straps become wider in the try-on image (\ie the green box of the first row) and the hands are unclear (\ie the red box of the third row). This happens because of the unawareness of the semantic layout and the relationship within the layout. ACGPN performs better than VTNFP, it can preserve hand detail but also failed to generate try-on details. It is because ACGPN uses TPS to warp clothes (mentioned in Section \ref{sub: IGMM}), and it uses a simple UNet to fuse the features, which makes it difficult to trade off characteristics of the target clothes and the reference person (mentioned in Section \ref{sub: TOFM}). Such as the sleeves of error lengths (\ie the green box of the third row) that caused by TPS \cite{duchon1977splines}, and the artifacts near the neck (\ie the red box of the fourth row, as the inner collar is not ignored) that caused by preserving too many characteristics of the warped clothes. 

However, AVTON does better both in preserving characteristics and trading off characteristics. Benefited from the LPM, it predicts limbs firstly and then provides limbs' information to the TOFM, it helps solve occlusion problems(\eg the blue box of the fourth row, the arm is clearer than others ). What's more, the IGMM warps clothes more reasonable to preserve styles and patterns (\eg the blue box of the third row, sleeves are the same length and patterns are clear), and the TOFM makes the try-on image more realistic due to characteristics' trade-off (\eg the yellow box of the fourth row, the inner collar should be ignored). In a word, AVTON can generate more realistic try-on images than VITON, CP-VTON, VTNFP, and ACGPN.

\begin{figure*}[h!]
\begin{center}
\includegraphics[width=0.9\textwidth]{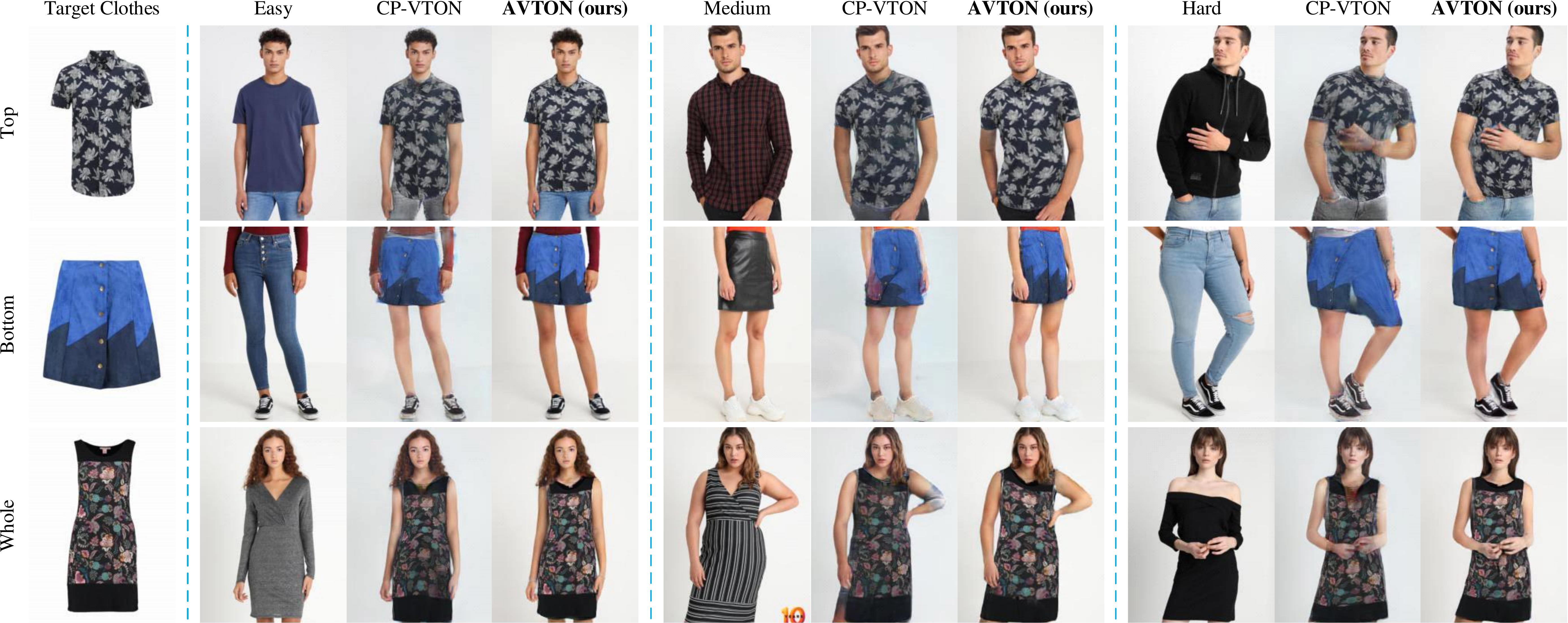}
\end{center}
   \captionof{figure}{\textbf{On the Zalando-Dataset}. Qualitative comparisons of CP-VTON and AVTON in different complexity levels. Our AVTON adapts successfully in cross-category try-on task and arbitrary shape and pose person.}
\label{fig:qualitative_results_2}
\end{figure*}

\textbf{On the Zalando-Dataset}. For a fair comparison, we retrained CP-VTON on Zalando-Dataset and selected the best-trained model. We put on different types of clothes for a person (Fig. \ref{fig:qualitative_results_1}) and put on one type of clothes for person of different complexity levels(Fig. \ref{fig:qualitative_results_2}). It is evident from the results that CP-VTON is not suitable for the all-type clothing try-on task, as CP-VTON uses TPS to warp clothes that we motioned in Section \ref{sub: IGMM}. And there are some defects such as the significant color difference of arm (Medium image of the third row in Fig. \ref{fig:qualitative_results_1}), as CP-VTON uses a single UNet to generate limbs in the try-on step that we motioned in Section \ref{sub: LPM}. However, AVTON can deal with these issues, it is benefited from the LPM, IGMM, and TOFM, where the LPM can predict the reasonable color of skin, the IGMM can warp clothes accurately, and the TOFM can make results more realistic.
\subsection{Quantitative Results}
We employ Structure SIMilarity (SSIM) \cite{wang2004image} to measure the similarity between try-on images and groundtruths, and Inception Score (IS) \cite{salimans2016improved} to measure the visual quality of try-on images. Specially, we measure SSIM in different complexity levels on the VITON-Dataset, and measure it in different types of clothes on the Zalando-Dataset.

\textbf{On the VITON-Dataset}. Table \ref{table 1} shows a quantitative comparison of our AVTON with VITON \cite{han2018viton}, CP-VTON\cite{wang2018toward}, VTNFP \cite{yu2019vtnfp} and ACGPN \cite{yang2020towards}. In our experiments, AVTON obtains a significant lead in all these metrics over the baseline methods. Specifically, in terms of SSIM, our method exceeds the best baseline method (\ie ACGPN) by \(0.010\), \(0.014\), and \(0.023\) for all complexity levels. And our method surpasses the best baseline method by \(0.195\) in terms of IS.

\begin{table}[!htp]\scriptsize
\centering
\begin{tabular}{c|cccc|c}
\toprule[1pt]
\multirow{2}{*}{Methods}    & \multicolumn{4}{c|}{SSIM}       & \multirow{2}{*}{IS} \\
\cline{2-5}
                            & Mean   & Easy   & Medium & Hard &                     \\
\midrule[1pt]
VITON                     & 0.783 & 0.787 & 0.779  & 0.779 & 2.650              \\
CP-VTON& 0.745& 0.753 &0.742& 0.729&2.757\\
VTNFP& 0.803&0.810&0.801& 0.788&2.784\\
ACGPN&0.845&0.854&0.841&0.828&2.829\\
AVTON (Vanilla) & 0.813 & 0.820 & 0.810  & 0.798 & 2.880        \\
AVTON (w/o \(\Psi\))         & 0.819 & 0.826 & 0.816  & 0.805 & 2.983       \\
AVTON (w/o LPM)             & 0.856 & 0.861 & 0.852  & 0.849 & 2.859        \\
AVTON (Full)                 & \textbf{0.859} & \textbf{0.864} & \textbf{0.855}  & \textbf{0.851} & \textbf{3.024}   \\
\bottomrule[1pt]           
\end{tabular}
\captionof{table}{\textbf{On the VITON-Dataset}. SSIM and IS results of five methods, while SSIM are measured in different complexity levels. AVTON (Vanilla), AVTON (w/o \(\Psi\)) and AVTON (w/o LPM) are for ablation study.}
\label{table 1}
\end{table}

\textbf{On the Zalando-Dataset}. As shown in Table \ref{table 2}, it presents a quantitative comparison of our AVTON with CP-VTON, where CP-VTON is retrained that mentioned in Section \ref{sub: qualitative results}. Compared to the SSIM of CP-VTON, AVTON increase it by \(0.053\), \(0.088\) and \(0.020\) for all-type clothes. And AVTON outperforms CP-VTON by \(0.420\) in terms of IS.

\begin{table}[!htp]\scriptsize
\centering
\begin{tabular}{c|cccc|c}
\toprule[1pt]
\multirow{2}{*}{Methods}    & \multicolumn{4}{c|}{SSIM}       & \multirow{2}{*}{IS} \\
\cline{2-5}
                            & Mean   & Top   & Bottom & Whole &                     \\
\midrule[1pt]
CP-VTON                     & 0.758 & 0.732 & 0.792  & 0.812 & 3.556            \\
AVTON (Vanilla) & 0.809 & 0.775 & 0.868  & 0.824 & 3.967        \\
AVTON (w/o \(\Psi\))         & 0.811 & 0.776 & 0.876  & 0.818 & 3.971       \\
AVTON (w/o LPM)             & 0.813 & 0.779 & 0.874  & 0.827 & \textbf{4.023}  \\
AVTON (Full)                 & \textbf{0.819} & \textbf{0.785} & \textbf{0.880}  & \textbf{0.832} & 3.976   \\
\bottomrule[1pt]           
\end{tabular}
\captionof{table}{\textbf{On the Zalando-Dataset}. SSIM and IS results of two methods, while SSIM are measured in different types of clothes. AVTON (Vanilla), AVTON (w/o \(\Psi\)) and AVTON (w/o LPM) are for ablation study.}
\label{table 2}
\end{table}
\subsection{Ablation Study}
\begin{figure}[h!]
\begin{center}
\includegraphics[width=1\linewidth]{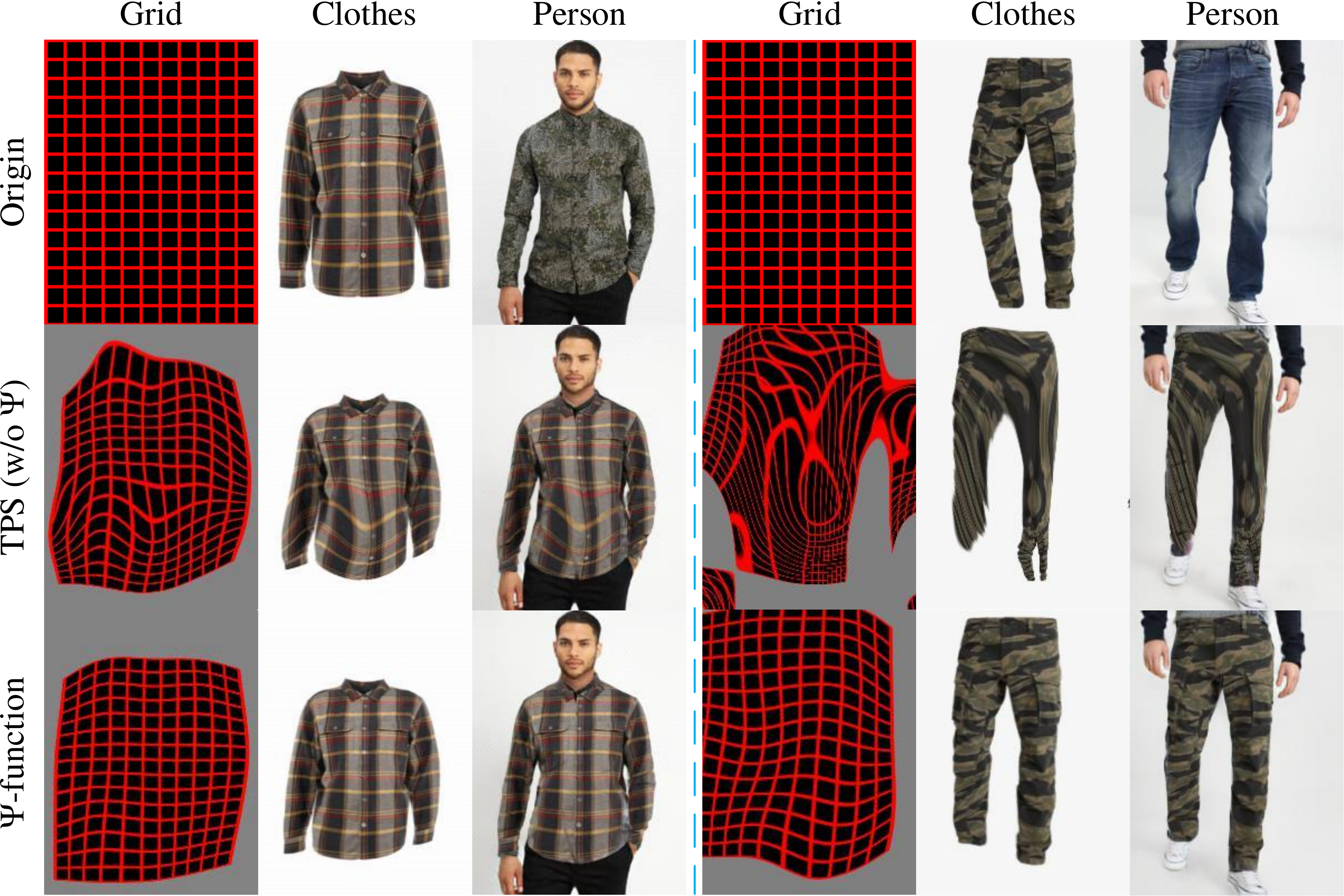}
\end{center}
   \captionof{figure}{Comparison of TPS and Wendland's \(\Psi\)-function.}
\label{fig:ablation_study_0}
\end{figure}

\begin{figure}[h!]
\begin{center}
\includegraphics[width=0.9\linewidth]{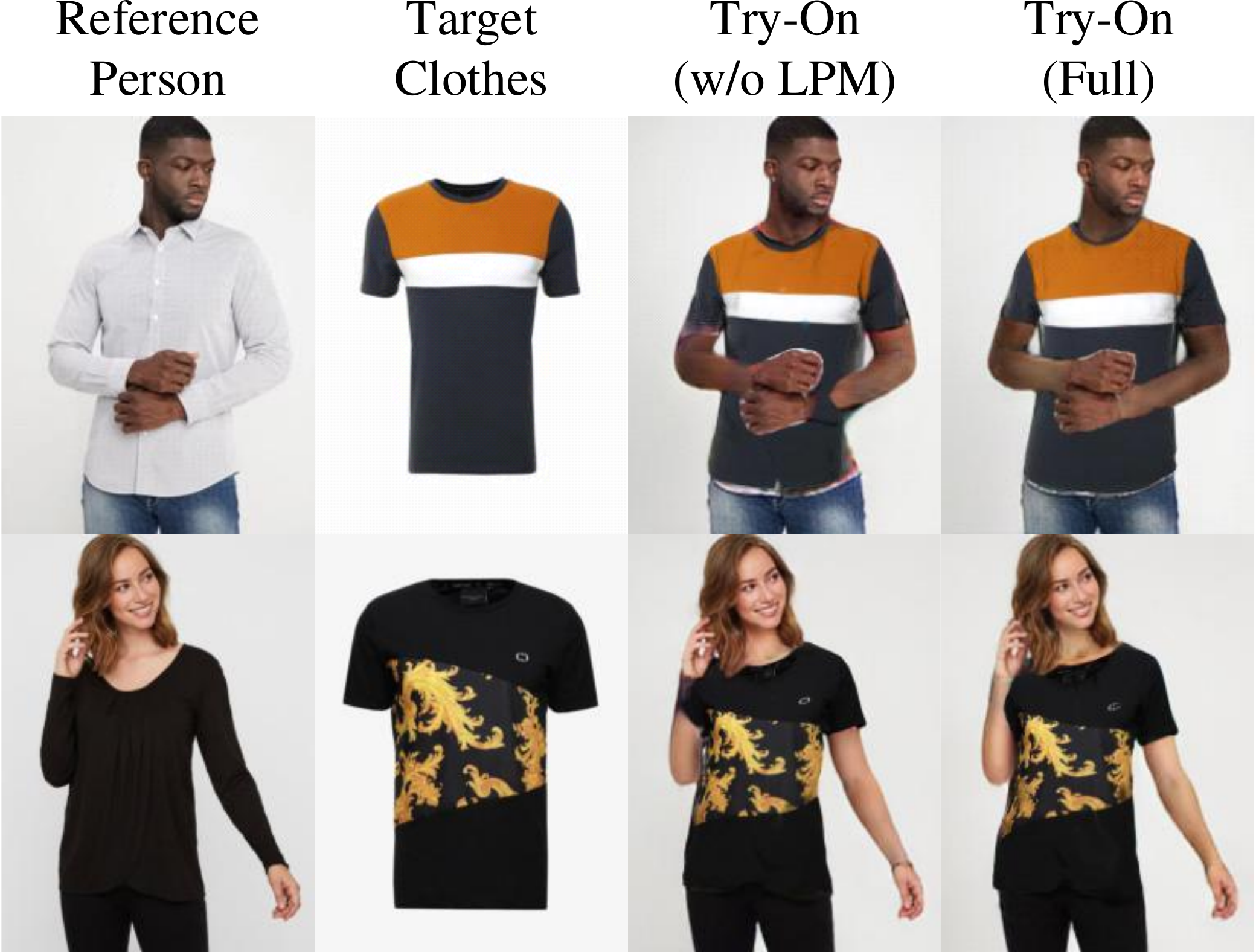}
\end{center}
   \captionof{figure}{Comparison of the results with and without the LPM.}
\label{fig:ablation_study_1}
\end{figure}

We evaluate the effectiveness of the LPM and \(\Psi\) (Wendland's \(\Psi\)-function). Similarly to the quantitative comparisons, we use SSIM and IS. As shown in Table \ref{table 1} and Table \ref{table 2}, Wendland's \(\Psi\)-function plays an important role, where AVTON (Full) surpasses the AVTON (w/o \(\Psi\)) by 0.046 and 0.008 in terms of the mean of SSIM, respectively. Here we show the visual comparison in Fig. \ref{fig:ablation_study_0}, it is obvious that Wendland's \(\Psi\)-function can warp clothes locally and smoothly, while TPS warped clothed globally that affects other parts.
Additionally, the LPM has a great impact on IS, compared to that of AVTON (w/o LPM), it is increased by 0.165 (Table \ref{table 1}). This of course increases IS score, as shown in Fig. \ref{fig:ablation_study_1}, the results without the LPM have broken arms. Especially in Table \ref{table 2}, the IS of AVTON (Full) is lower than AVTON (w/o LPM). It can be explained that the testing set contains Bottom and Whole cases, most of them do not have occlusion problems. Therefore, the prediction error caused by LPM can be avoided, and the IS of the results without the LPM is higher. To sum up, the LPM is necessary for the cross-category try-on task in our experiments.

\subsection{User Study}
As shown in Table \ref{table 3} and Table \ref{table 4}, we conduct two user studies on the VITON-Dataset and Zalando-Dataset. Among these two studies, we compare the ACGPN \cite{yang2020towards} and our proposed method AVTON on the VITON-Dataset from easy, medium, and hard cases, respectively, and we compare the CP-VTON \cite{wang2018toward} and AVTON on the Zalando-Dataset from the top, bottom, and whole cases, respectively. Note that we only choose the ACGPN as the baseline, as it is by far the best baseline in the VITON-Dataset based methods.  

\begin{table}[!htp]\scriptsize
\centering
\begin{tabular}{c|cccc}
\toprule[1pt]
                            Methods & Mean   & Easy   & Medium & Hard \\
\midrule[1pt]
ACGPN                     & 40.9\% & 39.0\% & 40.6\% & 43.0\%         \\
AVTON (Full)                 & \textbf{59.1\%} & \textbf{61.0\%} & \textbf{59.4\%}  & \textbf{57.0\%} \\
\bottomrule[1pt]           
\end{tabular}
\captionof{table}{\textbf{On the VITON-Dataset}. User study results between ACGPN and our proposed method AVTON in different complexity levels.}
\label{table 3}
\end{table}

Specifically, we invite 40 volunteers to complete the experiment. In each study, each volunteer is assigned 50 image pairs in a case and asked to select the most realistic image between two virtual try-on results. Both two studies show that the AVTON has better performance than other methods in all-type clothing and cross-category try-on tasks.

\begin{table}[!htp]\scriptsize
\centering
\begin{tabular}{c|cccc}
\toprule[1pt]
                            Methods & Mean   & Top   & Bottom & Whole \\
\midrule[1pt]
CP-VTON                     & 14.7\% & 14.1\% & 15.3\% & 14.6\%         \\
AVTON (Full)                 & \textbf{85.3\%} & \textbf{85.9\%} & \textbf{84.7\%}  & \textbf{85.4\%} \\
\bottomrule[1pt]           
\end{tabular}
\captionof{table}{\textbf{On the Zalando-Dataset}. User study results between CP-VTON and our proposed method AVTON in different types of clothes.}
\label{table 4}
\end{table}

\subsection{Arbitrary Clothing Collocation}
\begin{figure*}[h!]
\begin{center}
\includegraphics[width=1\textwidth]{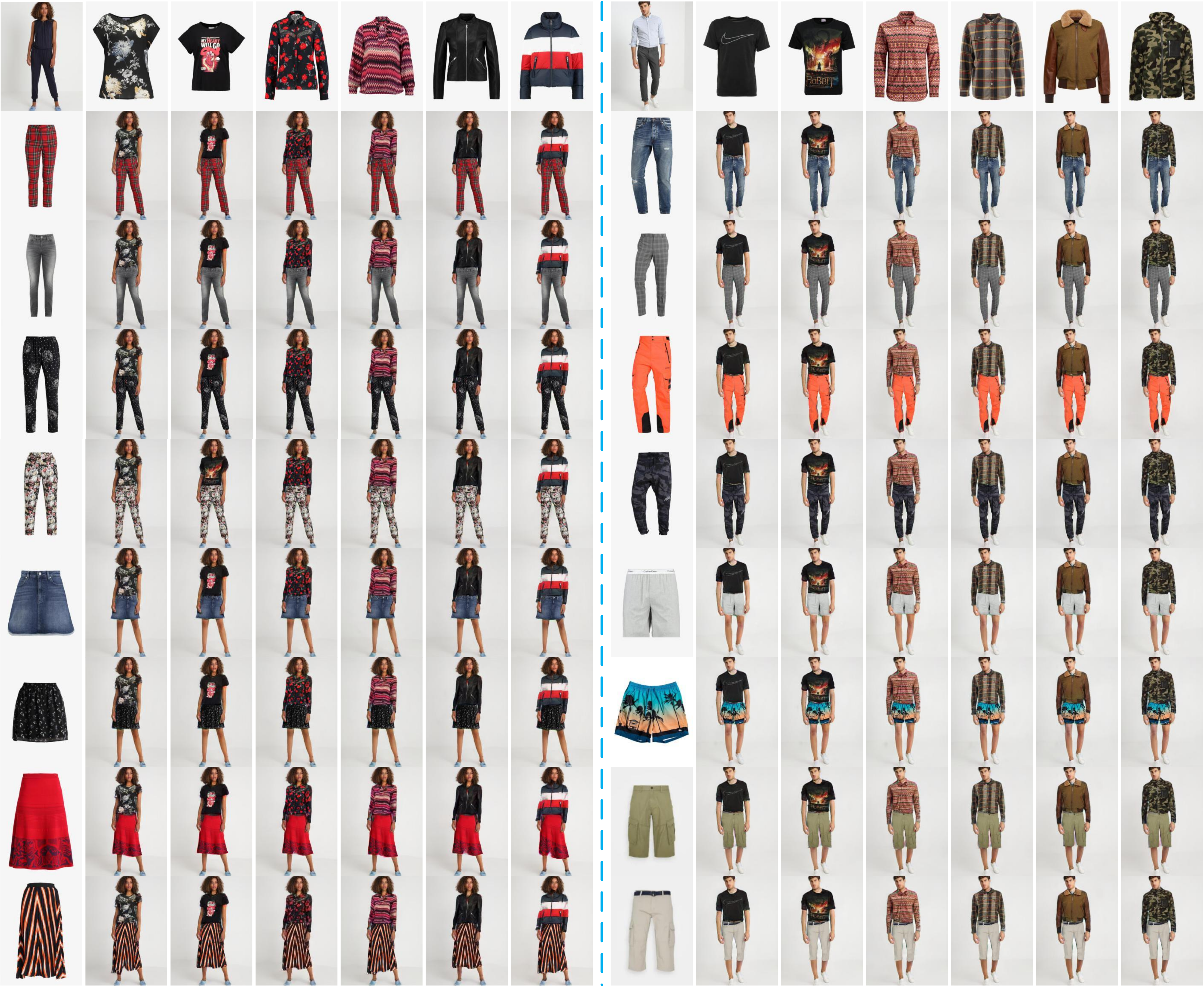}
\end{center}
   \captionof{figure}{The clothing collocation results of two parts. The upper left image of each part represents the reference person, the upper column of each part represents the target tops, and the left column of each part represents the target bottoms. It can be seen that our AVTON can match clothes arbitrarily.}
\label{fig:addition_results}
\end{figure*}

In real life, a single clothing virtual try-on cannot meet people's needs, and clothing collocation can improve people's favor of virtual try-on. Hence, We conduct an extra experiment to show the results of arbitrary clothing collocation (Fig. \ref{fig:addition_results}). During the experiment, we first try on tops with our AVTON and get the intermediate results, then try on bottoms based on the intermediate results, and finally get the clothing collocation results. It can be seen from the experimental results that due to the characteristics-preserving function of the LPM and IGMM, the target clothes and the reference person characteristics can still be retained after two try-on steps, and benefit from the characteristics trade-off function of TOFM, the final try-on images are natural and realistic.

\section{Conclusion}
In this paper, we propose a novel virtual try-on network, named AVTON, which aims at handling all-type clothing try-on task (tops, bottoms, and whole clothes) and cross-category try-on task (\eg long sleeves \(\leftrightarrow\) short sleeves or long pants \(\leftrightarrow\) skirts, \etc). Following extensive simulation results, we can observe: 1) the Limbs Prediction Module can well predict the human body parts by preserving the characteristics of the reference person. This is especially good for handling cross-category try-on task (\eg long sleeves \(\leftrightarrow\) short sleeves or long pants \(\leftrightarrow\) skirts, \etc), where the exposed arms or legs with the skin colors and details can be reasonably predicted; 2) the developed Improved Geometric Matching Module can achieve better warped performance than TPS based method, which can well characterize the clothes feature according to the geometry of target person. 3) while the Trade-Off Fusion Module (TOFM) can fuse both the information from render person as well as warped clothes to generate a realistic try-on image. Extensive simulations based on varied try-on tasks have been conducted. Simulation results based on Quantitative, qualitative evaluation and user study illustrate the great superiority of our AVTON over the state-of-the-art methods.

{\small
\bibliographystyle{ieee}
\bibliography{try_on_ref}
}

\begin{IEEEbiography}[{\includegraphics[width=1in,height=1.25in,clip,keepaspectratio]{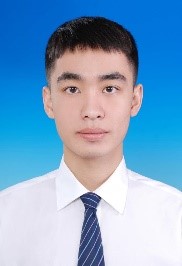}}]{Yu Liu} received the B.E degree from Donghua University in 2019. He is currently a Master candidate in Donghua University, Shanghai, P. R. China. His research interests include deep learning, pattern recognition, image synthesis.
\end{IEEEbiography}

\begin{IEEEbiography}[{\includegraphics[width=1in,height=1.25in,clip,keepaspectratio]{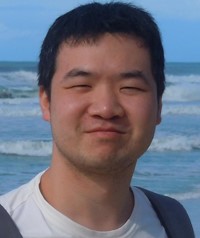}}]{Mingbo Zhao} (SM'19) received the Ph.D. degree in computer engineering from the Department of Electronic Engineering, City University of Hong Kong, Kowloon, Hong Kong S.A.R., in Jan. 2013. After that he also worked in City University of Hong Kong as a Postdoctoral researcher. He is currently a full professor in Donghua University, Shanghai, P. R. China. He has authored or co-authored over 50 technical papers published at prestigious international journals and conferences, including the IEEE TRANSACTIONS ON KNOWLEDGE AND DATA ENGINEERING, the IEEE TRANSACTIONS ON IMAGE PROCESSING, the IEEE TRANSACTIONS ON INDUSTRIAL INFORMATICS, the IEEE TRANSACTIONS ON INDUSTRIAL ELECTRONICS, the Pattern Recognition, Neural Networks, Knowledge based Systems, His current research interests include pattern recognition and Machine Learning. He is serving as an Associate Editor (AE) for IEEE ACCESS and Leader Guest Editor for Neural Computing and Application.
\end{IEEEbiography}

\begin{IEEEbiography}[{\includegraphics[width=1in,height=1.25in,clip,keepaspectratio]{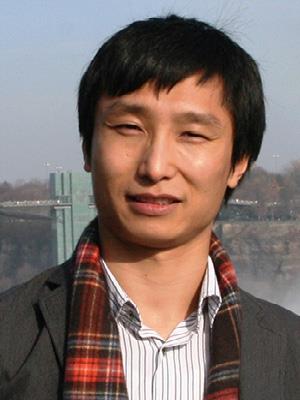}}]{Haijun Zhang} (SM'19) received the B.Eng. and master’s degrees from Northeastern University, Shenyang, China, and the Ph.D. degree from the Department of Electronic Engineering, City University of Hong Kong, Hong Kong, in 2004, 2007, and 2010, respectively. He was a Post-Doctoral Research Fellow with the Department of Electrical and Computer Engineering, University of Windsor, Windsor, ON, Canada, from 2010 to 2011. Since 2012, he has been with the Shenzhen Graduate School, Harbin Institute of Technology, Harbin, China, where he is currently a Full Professor of computer science. His current research interests include multimedia data mining, machine learning, and computational advertising. Dr. Zhang is also an Associate Editor of Neurocomputing, Neural Computing and Applications, and Pattern Analysis and Applications.
\end{IEEEbiography}

\begin{IEEEbiography}[{\includegraphics[width=1in,height=1.25in,clip,keepaspectratio]{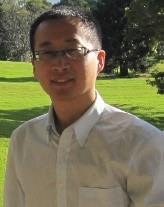}}]{Zhao Zhang} (SM'17) received the Ph.D. degree from the City University of Hong Kong, Hong Kong, in 2013. He was a Visiting Research Engineer with the National University of Singapore, Singapore, from February to May 2012. He visited the National Laboratory of Pattern Recognition, Chinese Academy of Sciences, from September to December 2012. From October 2013 to October 2018, he was an Associate Professor with Soochow University, Suzhou, China, where he was with the School of Computer Science and Technology. He is currently with the School of Computer and Information, Hefei University of Technology, Hefei, China. His current research interests include multimedia data mining and machine learning, image processing, and computer vision. Dr. Zhang is a member of the Association for Computing Machinery (ACM). He has been serving as a Senior PC Member or the Area Chair of ECAI, BMVC, PAKDD, and ICTAI, and a PC Member for more than ten popular prestigious conferences, including CVPR, ICCV, IJCAI, AAAI, ACM MM, ICDM, CIKM, and SDM. He has served or is serving as an Associate Editor (AE) for IEEE Trans. on Image Processing and Signal Processing.
\end{IEEEbiography}

\begin{IEEEbiography}[{\includegraphics[width=1in,height=1.25in,clip,keepaspectratio]{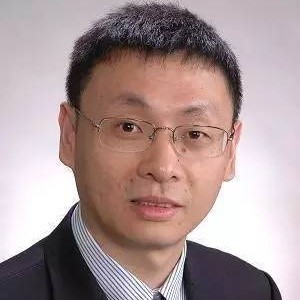}}]{Shuicheng Yan} (F'17) is currently the Vice President and the Chief Scientist with Qihoo 360 Technology Co., Ltd., Beijing, China, where he is the Head of the 360 Artificial Intelligence Institute. He is also a tenured Associate Professor with the National University of Singapore, Singapore. He has authored or coauthored about 500 high-quality technical articles, with Google Scholar citation over 25 000 times and the H-index of 70. His current research interests include computer vision, machine learning, and multimedia analysis. Dr. Yan is a fellow of the International Association for Pattern Recognition and the Distinguished Scientist of the Association for Computing Machinery (ACM). His team received seven times winner or honorable-mention prizes in five years over the PASCAL Visual Object Classes and the ImageNet Large Scale Visual Recognition Challenge competitions that are the core competitions in the field of computer vision, along with over ten times best (student) paper awards, especially a Grand Slam at the ACM Multimedia, the top conference in the field of multimedia, including the Best Paper Award, the Best Student Paper Award, and the Best Demo Award. He was a Thomson Reuters Highly Cited Researcher from 2014 to 2016.
\end{IEEEbiography}

\end{document}